\documentclass[11pt]{article}
\usepackage{amsmath,amssymb}
\usepackage{graphicx}
\usepackage{hyperref}
\usepackage{geometry}
\title{Addressing Heterogeneity in Federated Learning: Challenges and Solutions for a Shared Production Environment}
\author{
    Tatjana Legler\textsuperscript{1,2} \\
    \texttt{tatjana.legler@rptu.de}
    \and
    Vinit Hegiste\textsuperscript{1} \\
    \texttt{vinit.hegiste@rptu.de}
    \and
    Ahmed Anwar\textsuperscript{2} \\
    \texttt{ahmed.anwar@dfki.de}
    \and
    Martin Ruskowski\textsuperscript{1,2} \\
    \texttt{martin.ruskowski@dfki.de}
}
\date{\textsuperscript{1}University of Kaiserslautern-Landau (RPTU), Germany \\ 
\textsuperscript{2}German Research Center for Artificial Intelligence (DFKI), Germany}

\begin{document}
\maketitle
\begin{abstract}
Federated learning (FL) has emerged as a promising approach to training machine learning models across decentralized data sources while preserving data privacy, particularly in manufacturing and shared production environments. However, the presence of data heterogeneity variations in data distribution, quality, and volume across different or  clients and production sites, poses significant challenges to the effectiveness and efficiency of FL. This paper provides a comprehensive overview of heterogeneity in FL within the context of manufacturing, detailing the types and sources of heterogeneity, including non-independent and identically distributed (non-IID) data, unbalanced data, variable data quality, and statistical heterogeneity. We discuss the impact of these types of heterogeneity on model training and review current methodologies for mitigating their adverse effects. These methodologies include personalized and customized models, robust aggregation techniques, and client selection techniques. By synthesizing existing research and proposing new strategies, this paper aims to provide insight for effectively managing data heterogeneity in FL, enhancing model robustness, and ensuring fair and efficient training across diverse environments. Future research directions are also identified, highlighting the need for adaptive and scalable solutions to further improve the FL paradigm in the context of Industry 4.0.
\end{abstract}

\maketitle


\section{Introduction}
\label{intro}
Federated Learning (FL) is a collaborative learning approach that enables the training of models across multiple decentralized devices or servers holding local data samples, without exchanging their data \cite{Kairouz.2021}. This is achieved by training multiple clients on their local data, computing model updates, and then aggregating (e.g. averaging) them on a central server \cite{McMahan.2017}. By keeping data on local devices and only sharing model updates, FL minimizes the risk of data breaches and preserves user privacy. Techniques such as differential privacy and secure multi-party computation can be applied to further enhance security and privacy \cite{Mothukuri.2021}. In addition to the aspect of data sovereignty and privacy, employing an FL model circumvents the need for training a new model from scratch at each location, thereby enhancing energy efficiency \cite{Liu.2020}. Due to its inherent distributed approach, FL is also better able to scale and respond to the failures of individual participants, making it more robust \cite{Xu.2020}. Moreover, FL models are adept at generalizing across diverse scenarios, further underscoring their practical utility in distributed learning environments. Since the global server lacks information about the clients, heterogeneity such as varying data distributions must be addressed differently than in the centralized case, leading to new challenges and therefore new approaches to solving them.
Section 2 initially explores the various types of heterogeneity relevant in the context of FL. Subsequently, Section 3 identifies those that are particularly significant in the production environment and presents preliminary methods for addressing them.

\section{State of the Art}
First, some different types of heterogeneity are outlined briefly before exploring potential strategies to mitigate them, as these strategies frequently address multiple issues simultaneously. The literature lists various types that are not consistently defined or clearly differentiated, but they generally encompass the following categories \cite{Ye.2024,Abdelmoniem.2023}:

\textit{Device heterogeneity} in FL describes variations in computational power, network connectivity, and energy constraints or other hardware restrictions among participating clients \cite{Li.2024}. This type of heterogeneity challenges the uniform application of methods as clients range from powerful servers to resource-limited mobile and IoT devices. Clients with limited computational capabilities and energy constraints may not perform complex computations or frequent communications, impacting the overall learning process. Additionally, disparities in network connectivity can result in uneven data transmission rates, further complicating model synchronization and convergence \cite{Luo.2022}. System heterogeneity sometimes also includes model heterogeneity.

\textit{Model heterogeneity} occurs when clients use different model architectures, making it difficult to collaborate on a common model \cite{Alam.2022}. Traditional FL methods are limited to training models with the same structures, as the simple aggregation of weights is only feasible when each weight has a counterpart in other clients. This hinders applicability in scenarios with different hardware and communication networks, where otherwise the models could be selected according to the available computing power, e.g. smaller and efficient models on edge device and more powerful ones on high performance computers. To overcome this challenge, much more sophisticated approaches than simple averaging are required and often are based on knowledge distillation \cite{Huang.2022,Zhu.2021c}.

The term system heterogeneity often combines device and model heterogeneity. Both types of heterogeneity are crucial when considering mobile devices like smartphones, which operate across diverse hardware configurations supported by a single operating system. It is essential that none of these configurations are excluded in the optimization of a joint model. Similarly, in applications such as data collection from vehicles in preparation for autonomous driving, the hardware may vary significantly \cite{Fantauzzo.2022}. However, the diversity of this data becomes is even more critical to achieve a well generalized model \cite{Li.2022b}.

In dependence on the aforementioned types, \textit{participation heterogeneity} can occur, for example, an unstable internet connection can lead to some clients frequently joining and leaving the FL system, leading to irregular participation \cite{Pene.2024}. Depending on the volume of new data generated and the selection criteria for participation, some clients may not produce sufficient new data to qualify for participation in a communication round, resulting in a fluctuating frequency, that can affect the consistency and convergence of the global model.

\textit{Data Heterogeneity} refers to variability in data distribution across clients that can lead to biases and affect model performance \cite{Yang.2021c}.
Independent and Identically Distributed (IID) variables refer to a sequence of random variables that are statistically independent and follow the same underlying probability distribution \cite{Jacobs.1992}. The IID assumption is fundamental in probability theory and statistics, facilitating the modeling of numerous real-world phenomena, such as repeated trials of an experiment or the behavior of a system over time. Although these assumptions are crucial for constructing and validating statistical models, real-world problems seldom exhibit true uniform distribution \cite{Hsieh.2020}.
However, the definition of non-IID is more varied as there are different ways of deviating from the uniform distribution \cite{Kairouz.2021}.


\begin{figure}[t]\vspace*{4pt}
\centerline{\includegraphics[width=1.0\textwidth]{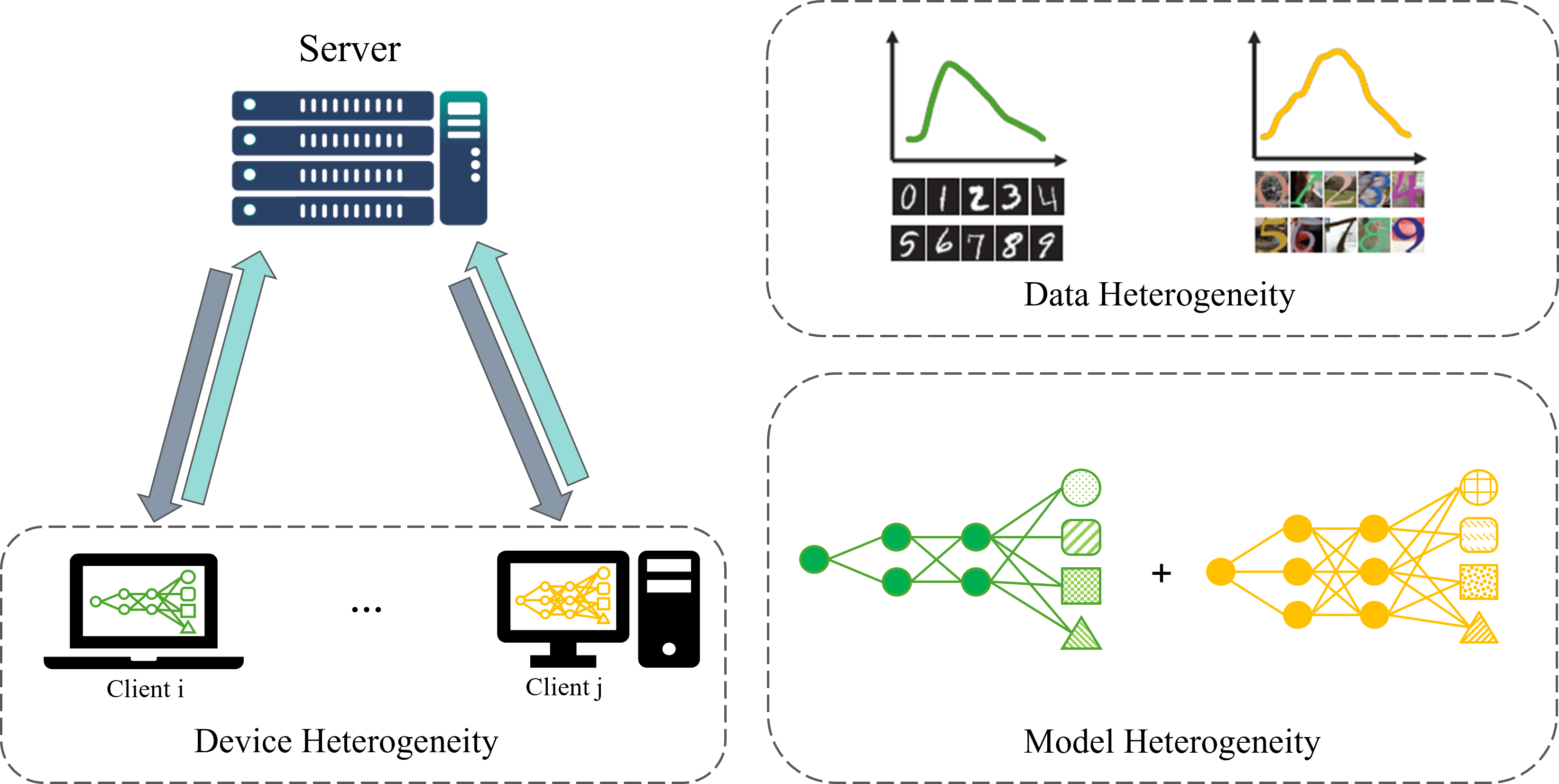}}
\caption{Overview of heterogeneity in Federated Learning systems. The figure illustrates three types of heterogeneity that can affect FL systems: device heterogeneity (differences in computational resources among clients), data heterogeneity (variations in data distributions across clients), and model heterogeneity (differences in model architectures or parameters used by different clients). In parts based on \cite{Huang.2022}} 
\label{fig:fl_het}
\end{figure}

There are different types of shifting the relationship between the input (features) and output data (labels) \cite{Y.2019}:
\begin{itemize}
\item \textit{Covariate shift}: Feature distribution skew 
\item \textit{Prior probability shift}: Label distribution skew 
\item \textit{Concept shift}: Same label, different features or vice versa.
\end{itemize}

 Systems that are actively deployed may encounter temporal shifts in newly generated data over time, diverging from the original training dataset, a phenomenon known as \textit{dataset shift} \cite{Ovadia.2019}. Therefore, considerations for continual learning become essential, such as incorporating strategies to learn new information while retaining previously acquired knowledge (i.e., preventing catastrophic forgetting) \cite{Parisi.2019,Huang.2022}.

Additionally, there might be significant disparities in the volume of data that each client contributes to the global model, often referred to as \textit{quantity skew} or "unbalancedness". Data availability can vary greatly among clients, especially when initializing the system; some may have a large amount of historical data, while others may just have started with data acquisition. Factors previously mentioned concerning participation also influence data contribution.
Such imbalances can result in a bias in model training, as clients with larger datasets or multiple clients sharing a common dataset can disproportionately affect the global model. Additionally, such imbalances can cause convergence problems, as training on uneven data distributions may slow convergence or produce suboptimal outcomes, with the global model potentially overfitting to the more data-rich clients. Figure \ref{fig:fl_het} illustrates different types of heterogeneity.

\section{Addressing Heterogeneity in a Production Environment}
As described in preliminary work, we are looking at a shared production scenario in which companies can offer and request services \cite{Simon.2023}. These services can include hardware services such as milling or drilling of parts, as well as software services like quality inspection solutions as shown in \cite{Hegiste.2022}. While we assume that security aspects are covered through the use of a common platform, and none of the participants have malicious ulterior motives (e.g. planning data poisoning attacks), all the usual requirements for FL, such as not sharing any production data, still apply \cite{Legler.2023}. 
Our assumptions correspond to a cross-silo setting, in which a small number of clients, e.g. from different organizations, tend to participate \cite{HuangChao.2022}. In contrast to cross-device, where system and model heterogeneity is also important, organizations or companies are likely to have sufficient computing resources and stable network connections. We therefore focus on data heterogeneity, e.g., handling highly non-i.i.d. data across different clients.
The quality and statistical properties of data can significantly vary across clients due to diverse data collection methods, environments, or noise levels. Issues such as noisy data, which may contain errors, irrelevant information, or noise, can degrade model performance. Incomplete data, characterized by missing values or incomplete records, impacts the training process and the resultant model quality. Outliers or extreme values in the data can skew model training, particularly if they are not adequately addressed. Additionally, statistical heterogeneity manifests through variations in the mean and variance of features across clients, posing challenges in data standardization. Differences in correlation structures among different clients can further complicate the model's ability to generalize effectively across diverse data landscapes.

\begin{figure}
    \centering
    \includegraphics[width=1\linewidth]{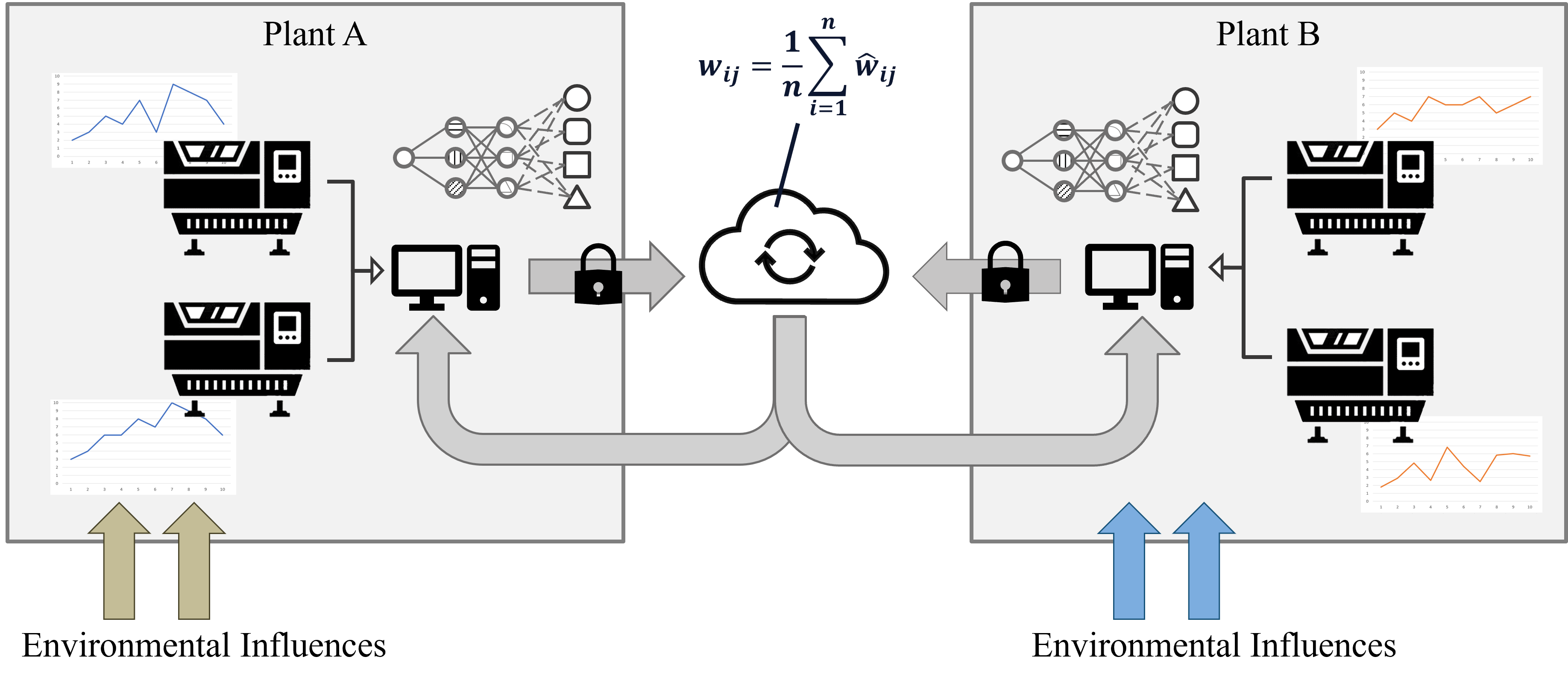}
    \caption{A federated learning system spanning multiple production plants. Only an aggregated model is shared for each plant, therefore enhancing data privacy.}
    \label{fig:fl_plants}
\end{figure}

A tiered (also: hierarchical or grouped) training approach, where clients with similar system capabilities will be grouped and perform a model aggregation within this group before a common model is sent to the next tier or global server \cite{Chai.2020,Chang.2023}.
This can make particular sense if a company has many of its own machines or devices that already benefit from the participation of a (local) FL system. Let's assume, for example, that a machine manufacturer supplies a customer with 100 machines, 20 of which are located in a hall. Although the machines are identical in construction and possibly produce the same part, there are always small deviations due to tolerances. This group of 20 machines will have similar environmental conditions and will probably also be equipped and maintained with the same or similar resources. Creating an FL system on their data alone will lead to a somewhat more stable and better-generalized model, but integrating the other groups will bring in more data diversity and thus unlock more of the potential. Even more of the benefits of FL can be leveraged when the next step is to go beyond corporate boundaries. Many companies keep their production data heavily protected, including how many units of each product have been produced in a given time. The mere possibility of disclosing the number of machines to a competitor can be an exclusion criterion for participation in a system.      
In such a case, however, the company can implement FL at least within its own locations. Given that data protection regulations across national borders also affect information flow within a company, this allows various sites to be interconnected (see Figure \ref{fig:fl_plants}). In a tier-based approach, the number of machines does not need to correlate with the number of FL clients, since even a single model update can be sent to the next higher tier. This has the advantage of disclosing even less information than traditional approaches when participating in the FL system. However, a disadvantage is that the central servers are unaware of the weighting of these model updates and may need to rely on alternative client-selection techniques (e.g., performance-/loss-based).

A more robust aggregation can be achieved by modifying the standard FedAvg algorithm to account for data heterogeneity, such as by weighting updates based on data size or quality. For example, \cite{Li.2023} propose Centered Kernel Alignment (CKA) to compute the similarity of feature maps in the output layer and enables fast model aggregation and improves global model accuracy in non-IID scenario .

The initial approach to client selection proposed in \cite{McMahan.2017} was to sample a random fraction of clients for the next communication round. Current client selection techniques investigate which parameters can be used to determine the next participants. Rather than clustering clients into tiers as previously described, clustering approaches can also be employed to temporarily identify clusters and use them to select a client from each. This method ensures a more comprehensive coverage of the entire spectrum.
Other approaches utilize training metrics such as local accuracy and training loss to identify and select the worst-performing clients, as these clients have the greatest potential for improvement and can therefore add the most to the global model. The concept of Contribution-Based Selection refers to methods that utilize the impact of a client on the global model as a criterion for selecting clients in subsequent communication rounds. In their work, \cite{Qiao.2023} employ the Shapley value, a concept from cooperative game theory, to estimate each client’s contribution to the global model. Additionally, \cite{Lin.2022} propose an advancement in contribution estimation by considering both gradient space and data space for individual clients. Generally speaking, the less information that needs to be shared in manufacturing, the better. Therefore, processes that operate solely based on weights or weight changes are preferred.

As the production of defects is highly costly, every company strives to eliminate them as effectively as possible. This naturally results in a significant discrepancy in the dataset, as data with good parts is significantly more likely than data with errors. To counteract this class imbalance, various techniques can be used, especially locally at the client. These include artificially enlarging the data set and changing the weighting of the classes. Synthetic data generation can be used particularly in use cases where there is a high imbalance or where it would be very costly to collect new real data \cite{Figueira.2022,Hegiste.2024}. It can be used to either fill in gaps in the real dataset, by creating new data points \cite{Jaipuria.2020} or augment the existing dataset \cite{Langevin.2022} to improve generalization. Furthermore, a local resampling of sparse data points can also mitigate the imbalance \cite{Khosla.2020}.

\section{Conclusion}
In conclusion, the exploration of heterogeneity in FL within a shared production environment has unveiled a complex landscape of challenges and potential solutions that could significantly advance the field of decentralized machine learning. Our review highlights the crucial need for robust, adaptive methods that can accommodate the unique constraints and characteristics of each client in the federated network. Strategies such as personalized modeling, advanced aggregation techniques, and thoughtful client selection have shown promise in mitigating the adverse effects of heterogeneity, thereby enhancing model performance and fairness across various settings.

Furthermore, the discussion emphasizes the importance of continued research into scalable, flexible solutions that can handle the dynamic and evolving nature of data and system architectures in real-world applications. By fostering a deeper understanding of these issues and continuously innovating on the solutions, the full potential of FL in industrial and commercial applications can be realized.

In overcoming these barriers, FL will not only improve model accuracy and training efficiency but will also pave the way for more secure, privacy-preserving, and collaborative machine learning endeavors in globally distributed networks. This study sets the stage for future research directions, urging a sustained commitment to addressing these challenges within the landscape of Industrie4.0 and beyond.
\section*{Acknowledgements}

This work was funded by the Carl Zeiss Stiftung, Germany under the Sustainable Embedded AI project (P2021-02-009).

\bibliographystyle{elsarticle-num.bst}
\bibliography{lit.bib}

\end{document}